\documentclass{article}

\PassOptionsToPackage{numbers, compress}{natbib}
\usepackage{cite}

\usepackage[preprint]{neurips_2025}

\usepackage[utf8]{inputenc} 
\usepackage[T1]{fontenc}    
\usepackage{hyperref}       
\usepackage{url}            
\usepackage{booktabs}       
\usepackage{amsfonts}       
\usepackage{nicefrac}       
\usepackage{microtype}      
\usepackage{xcolor}         

\usepackage{comment}
\usepackage{multirow}
\usepackage{algorithm,algpseudocode}
\usepackage{url}
\usepackage{graphicx} 
\usepackage{enumitem}
\usepackage{wrapfig}
\usepackage{subcaption}

\usepackage{chngpage}
\usepackage{array}
\usepackage{amsmath}
\usepackage{bbding}
\usepackage{verbatim}
\usepackage{wasysym} 
\usepackage{xspace}
\usepackage{tabularx}
\usepackage[outercaption]{sidecap}    
\usepackage{dsfont}

\usepackage{booktabs}  
\usepackage{circledsteps}

\usepackage{color, colortbl}
\definecolor{mygray}{gray}{0.95}
\definecolor{myred}{rgb}{1.0, 0.0, 0.0}

\usepackage{caption} 
\usepackage{amssymb}
\usepackage{pifont}
\makeatletter
\DeclareRobustCommand\onedot{\futurelet\@let@token\@onedot}
\def\@onedot{\ifx\@let@token.\else.\null\fi\xspace}

\makeatother

\title{Segment Any 3D-Part in a Scene from a Sentence}

\author{Hongyu Wu$^{1}$~\qquad Pengwan Yang$^{1,}$\footnotemark[2]~\qquad Yuki M. Asano$^{1,2}$~\qquad Cees G. M. Snoek$^{1}$\\
$^{1}$University of Amsterdam~\qquad $^{2}$Technical University of Nuremberg\\
Project Webpage : \href{https://3dpartseg.github.io/}{https://3dpartseg.github.io/} \\
}

\begin{document}

\renewcommand{\thefootnote}{\fnsymbol{footnote}}
\footnotetext[2]{Corresponding author}

 \begin{center}
\maketitle
\begin{figure}[h]
\centering
    \vspace{-15px}
    \captionsetup{type=figure}
    \includegraphics[width=0.99\linewidth]{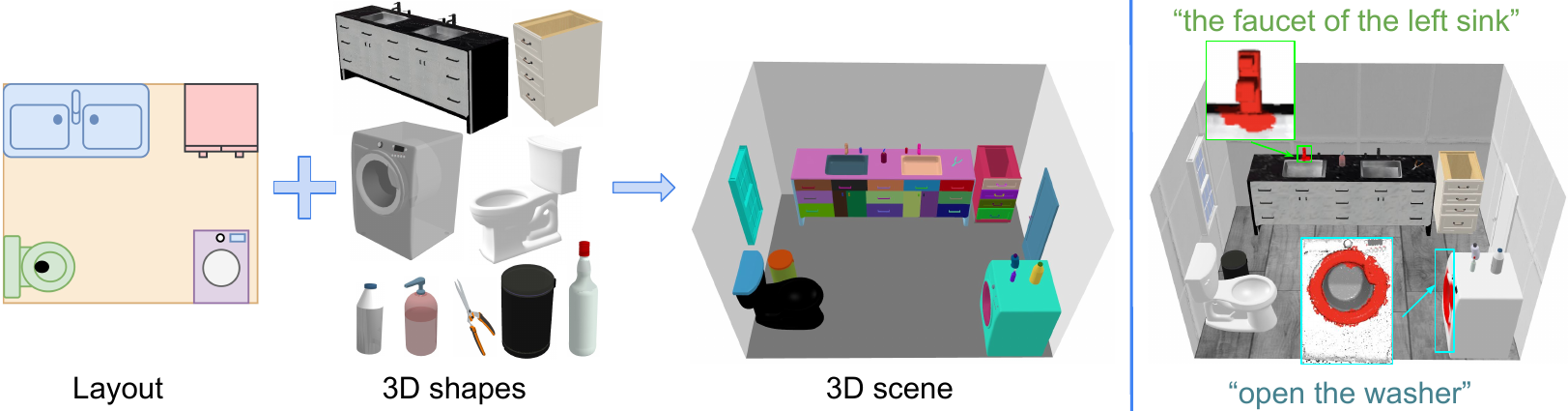}
    
    \caption{
    \textbf{The 3D-part scene understanding dataset and method.} 
     We propose the first large-scale 3D scene dataset with dense part annotations, based on an innovative approach for constructing 3D scenes with detailed part annotations (left). Using this new dataset, we introduce the 3D-part understanding task and method that enables flexible part segmentation and identification in s scene based on any sentence query (right).
     }
     \label{fig:homepage}
     \vspace{-5px}
\end{figure}
\end{center}
 \begin{abstract}
\vspace{-5pt}

This paper aims to achieve the segmentation of any 3D part in a scene based on natural language descriptions, extending beyond traditional object-level 3D scene understanding and addressing both data and methodological challenges. Due to the expensive acquisition and annotation burden, existing datasets and methods are predominantly limited to object-level comprehension. To overcome the limitations of data and annotation availability, we introduce the 3D-PU dataset, the first large-scale 3D dataset with dense part annotations, created through an innovative and cost-effective method for constructing synthetic 3D scenes with fine-grained part-level annotations, paving the way for advanced 3D-part scene understanding. On the methodological side, we propose OpenPart3D, a 3D-input-only framework to effectively tackle the challenges of part-level segmentation. Extensive experiments demonstrate the superiority of our approach in open-vocabulary 3D scene understanding tasks at the part level, with strong generalization capabilities across various 3D scene datasets.

\vspace{-5pt}
\end{abstract}  
 \section{Introduction}
\label{sec:intro}
\vspace{-5px}

Significant progress has been made in open-vocabulary 3D scene understanding, propelled by vision-language models trained on large-scale datasets~\citep{chen2023clip2scene, ding2023pla, ding2024lowis3d, peng2023openscene, rozenberszki2022language, yang2024regionplc, zeng2023clip2, takmaz2023openmask3d, nguyen2024open3dis, huang2025openins3d}. For instance, OpenMask3D~\cite{takmaz2023openmask3d} builds on Mask3D~\cite{schult2023mask3d} to generate 3D instance proposals, leveraging the CLIP model~\cite{radford2021learning} to align objects with a text query. Similarly, OpenIns3D~\cite{huang2025openins3d} incorporates text comprehension and object detection by adapting a 2D open-world model to 3D scenes. 
Currently, most open-vocabulary 3D scene understanding efforts are largely object-focused due to the availability of relevant data.
Recently, Search3D~\cite{takmaz2025search3d} tries to extend beyond object segmentation by constructing 2D object-part representations that correspond to 3D objects, relying on well-aligned RGB-D images. 

In this paper, we aim to advance open-vocabulary 3D scene understanding to the part level, enabling fine-grained segmentation of individual 3D parts within scenes. This task introduces new technical challenges and addresses the limitations of existing datasets, which are primarily designed for object-level understanding~\cite{dai2017scannet,replica19arxiv,yeshwanthliu2023scannetpp,dehghan2021arkitscenes,Matterport3D}. 
Recent datasets, such as MultiScan~\cite{mao2022multiscan} and SceneFun3D~\cite{delitzas2024scenefun3d}, have started to provide part-level annotations for 3D scenes, with MultiScan offering 5K
annotations across five categories in 117 scenes, and SceneFun3D delivering 14K
annotations with nine affordance labels across 710 scenes. While these are valuable contributions, they are constrained by the limitations of manual labeling, resulting in a relatively coarse granularity and an insufficient number of part annotations for training or fine-tuning robust part-segmentation models. 

To overcome these limitations, we propose a novel, cost-effective approach for constructing 3D scene data with detailed, fine-grained part-level annotations, bypassing the need for extensive manual scanning and labeling. As illustrated in Figure~\ref{fig:homepage}~(left), we create 3D scenes by combining annotated 3D shapes guided by structured layouts. This approach leads to the development of the 3D-PU (3D Part Understanding) dataset, a large-scale 3D dataset with 843,654 dense part annotations across 206 categories for 10,000 scenes, establishing a robust foundation for advancing part-level 3D scene understanding.

Building upon the part-level annotated scene dataset, we introduce the task of open-vocabulary 3D part segmentation, enabling the segmentation of any 3D part within a scene by a text query, as shown in Figure~\ref{fig:homepage}~(right). To address this task, we propose OpenPart3D, a 3D-input-only framework designed to efficiently and accurately segment 3D parts in complex scenes. As illustrated in Figure~\ref{fig:method}, the \textit{Room-Tour Snap Module} generates multiple view images from the 3D scene by positioning cameras at predefined locations with optimized poses, ensuring that small objects and parts are clearly visible in the captured views. These images are then input into a 2D open-vocabulary model to generate 2D part masks relevant to the text query. The \textit{View-Weighted 3D-Part Grouping Module} leverages these 2D part masks across multiple views, incorporating varying weights, to extract geometrically coherent regions from the scene point cloud and aggregate them into precise 3D parts.

Overall, we address both the data limitations and technical challenges inherent in the difficult task of open-vocabulary part segmentation in complex 3D scenes, achieving superior performance.
\begin{itemize}
\vspace{-5px}
\item To circumvent the high expense of manual 3D scanning and labeling, we propose a cost-effective approach for constructing 3D scene data with detailed, fine-grained part annotations. We present the 3D-PU dataset, the first large-scale 3D scene dataset with comprehensive part annotations.

\item On the technical side, we propose OpenPart3D, a 3D-input-only framework designed to efficiently and accurately segment 3D parts in complex scenes. 

\item Our proposed approach demonstrates superior performance over baselines in open-vocabulary 3D part segmentation tasks, while exhibiting robust generalization across various 3D scene datasets.
\end{itemize}
\vspace{-5px}

 \section{Related work}
\label{sec:related}
\vspace{-5px}

\noindent \textbf{Open-vocabulary 3D scene understanding.} Open-vocabulary understanding of 3D scenes has recently garnered considerable attention~\cite{chen2023clip2scene,ding2023pla,ding2024lowis3d,peng2023openscene,rozenberszki2022language,yang2024regionplc,zeng2023clip2,takmaz2023openmask3d,nguyen2024open3dis,huang2025openins3d}. OpenScene~\cite{peng2023openscene} pioneered open-vocabulary 3D scene understanding by leveraging per-pixel image features extracted from posed images of a scene to obtain a task-agnostic, point-wise scene representation. OpenMask3D~\cite{takmaz2023openmask3d} employs well-aligned 2D images to leverage 2D image features for 3D instance mask proposals, achieving strong open-vocabulary instance segmentation. OpenIns3D~\cite{huang2025openins3d} introduced the ``Snap and Lookup'' strategy, which has proven effective as a 3D object recognition engine, delivering competitive open-vocabulary instance segmentation without relying on well-aligned images.
Currently, most existing open-vocabulary 3D scene understanding approaches focus on object-level semantic segmentation or instance segmentation. 
More recently, Search3D~\cite{takmaz2025search3d} has sought to extend beyond object segmentation by constructing 2D image object-part representations that correspond to 3D objects, thereby enabling hierarchical object-part segmentation. However, a key limitation of Search3D lies in its reliance on well-aligned 2D-3D input pairs, which are often unavailable or impractical in real-world applications. 
Building on the spirit of Search3D, we aim to advance open-vocabulary 3D scene understanding to the part-level, extending the capabilities of object-level understanding and enabling query-based part segmentation directly within 3D scenes. Importantly, we simplify input requirements to 3D data only, enhancing method flexibility and compatibility across diverse settings.

\noindent \textbf{3D scene datasets.} Existing 3D scene datasets are constrained by their reliance on multiple observations of the same scene and lack extensive part-level information. Datasets like ScanNet~\cite{dai2017scannet}, Replica~\cite{replica19arxiv}, and ScanNet++~\cite{yeshwanthliu2023scannetpp} focus on single-room reconstructions, providing dense semantic object annotations on mesh vertices, while ARKitScenes\cite{dehghan2021arkitscenes}—the largest room-scale 3D dataset—includes 1,661 unique scenes but only offers 3D object bounding box annotations. Matterport3D~\cite{Matterport3D}, a building-scale dataset, offers manual object-level semantic annotations. Recently, two datasets have provided part-level annotations for 3D scenes. MultiScan~\cite{mao2022multiscan} includes 5,129 part annotations across 5 categories for 117 scenes, and SceneFun3D~\cite{delitzas2024scenefun3d} delivers 14,867 part-level annotations with 9 affordance labels across 710 scenes. These are valuable contributions, however, as they are limited by the constraints of manual labeling, the part annotation granularity remains coarse and the relatively small number of part annotations is insufficient for training or fine-tuning robust part-segmentation models.
To address these limitations, we propose a novel approach for constructing 3D scene data with extensive fine-grained part-level annotations at a low cost. We introduce the first large-scale 3D scene dataset with detailed part annotations, named the 3D-PU dataset. The 3D-PU dataset includes 843,654 dense part annotations across 206 categories for 10,000 scenes, creating a robust foundation for advancing 3D part-level scene understanding.

\noindent \textbf{Vision-language models for 3D.}
The recent success of large-scale model training has led to the development of a range of vision-language models (VLMs). Models such as CLIP~\cite{radford2021learning} provide a joint embedding space for both image and text, typically taking a single image as input and generating a global image embedding. While effective for tasks like image classification, these representations are limited in their ability to handle tasks requiring spatial localization. SAM~\cite{kirillov2023segment} and SAM2~\cite{ravi2024sam2} provide the ability to generate class-agnostic masks for any 2D instances, while Mask3D~\cite{schult2023mask3d} extends this capability to 3D scenes, predicting class-agnostic masks for all 3D objects.
To address localization-based open-vocabulary detection and segmentation, a series of methods~\cite{xiao2024florence,liu2023grounding} have been proposed. Florence2~\cite{xiao2024florence} and GroundingDino~\cite{liu2023grounding} feature pixel-aligned embeddings, associating each pixel with an embedding vector in the vision-language space, enabling flexible part querying on 2D planes.
Leveraging the spatial capabilities of VLMs, we propose extending these models to the novel task of open-vocabulary 3D part segmentation in scenes, thus minimizing the need for extensive retraining.
\vspace{-5px}

 \section{3D-PU dataset}
\label{sec:dataset}
\vspace{-5px}

\begin{wrapfigure}{r}{0.5\textwidth}
    \centering
    \vspace{-10px}
     \includegraphics[width=0.49\textwidth]{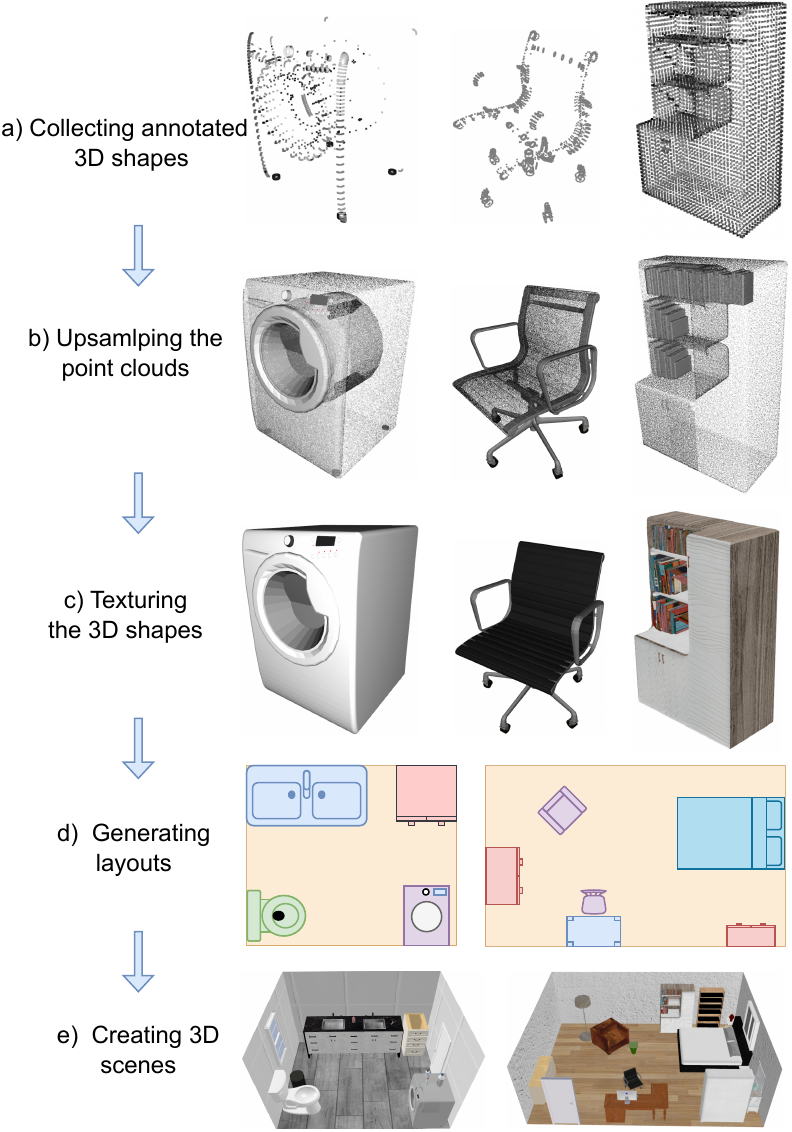}
	\caption{\textbf{3D-PU dataset construction} consists of the following core steps: a) collecting a large volume of 3D shapes with part annotations from PartNet~\cite{mo2019partnet}, SAPIEN~\cite{Xiang2020sapien}, and ShapeNetPart~\cite{chang2015shapenet}, b) uniformly upsampling the point clouds from meshes~\cite{yuksel2015sample}, c) generating textures and colors for each 3D shape using both an offline text-guided texturing model~\cite{richardson2023texture} and an online texturing service, d) generating diverse layouts with the assistance of ChatGPT-4, and e) constructing 3D scenes by populating part-annotated objects into the defined scene layouts.
    }
	\label{fig:construction}
    \vspace{-8px}
\end{wrapfigure}

A straightforward approach to building a part-annotated 3D scene dataset is to manually label all parts in 3D scenes; however, this is prohibitively costly and impractical. Another common method involves projecting 2D pseudo-part masks from 2D foundation models like SAM2~\cite{ravi2024sam2} onto 3D scenes, but this technique falls short of providing precise and comprehensive small-part annotations. In contrast to these methods, we propose constructing 3D scenes with part annotations by leveraging existing 3D shapes that already contain part-level annotations. This approach allows for more accurate and scalable dataset creation. Figure~\ref{fig:construction} outlines the key steps in the process.

\noindent\textbf{Collecting annotated 3D shapes.} We gather 3D shapes with part annotations from PartNet~\cite{mo2019partnet}, SAPIEN~\cite{Xiang2020sapien}, and ShapeNetPart, a subset of ShapeNet~\cite{chang2015shapenet}. Each shape includes a point cloud, mesh, and fine-grained part annotations. In total, we collected 51,300 distinct 3D shapes across 85 object categories, along with 587,600 part annotations across 206 part classes. 

\noindent\textbf{Upsampling the point clouds.} As shown in Figure~\ref{fig:construction}~a), the original collected shapes typically have denser points on salient parts and sparser points on inner or flat areas. To create uniformly distributed and denser point clouds, we upsample the points of each shape from its mesh~\cite{yuksel2015sample}. Shapes are categorized into large, medium, and small based on their object categories. Large shapes are upsampled to 102,400 points, medium shapes to 51,200 points, and small shapes to 25,600 points. This upsampling results in denser and more complete point clouds. 

\noindent\textbf{Texturing the 3D Shapes.} To enhance the realism of the 3D scenes, we generate textures and color specifications for each 3D shape. Initially, we use GPT-4 to generate ten natural language prompts, each containing stylized descriptors and color specifications. These ten prompts, along with the corresponding 3D shapes, are processed by both an offline text-guided texturing model~\cite{richardson2023texture} and an online texturing service provided by meshy.ai, yielding a total of 20 textured variations. Subsequently, we compute the similarity score between each textured shape and its corresponding text prompt in the language-vision embedding spaces of CLIP2~\cite{zeng2023clip2} and Uni3D~\cite{zhou2023uni3d}. Finally, we select the five textured shapes with the highest similarity scores from the pool of 20. 

\noindent\textbf{Generating layouts.} We use layouts to guide the structured generation of 3D scenes. With the assistance of ChatGPT-4, we designed 100 distinct layouts representing a wide range of scenes, such as bedrooms, bathrooms, kitchens, living rooms, offices, classrooms, storage rooms, and more. These automatically generated layouts were manually reviewed to correct issues such as object collisions, incorrect orientations, and other inconsistencies. As shown in Figure~\ref{fig:construction}~d), each layout specifies the required large objects and their placement within the scene.

\noindent\textbf{Creating 3D scenes.} Guided by the predefined layouts, we position the specified large 3D objects in their designated locations. Next, we add randomly selected small objects at various positions within the scene to enhance realism and diversity. This process results in natural, realistic 3D scenes with comprehensive part annotations.

\noindent\textbf{Generating text queries.} In the 3D scenes, each part annotation is assigned an ``object\_part'' label, such as ``door\_handle'' or ``table\_leg.'' These "object\_part" labels can serve directly as text queries for open-vocabulary part segmentation. In addition to direct text queries, we generate descriptive queries from the ``object\_part'' labels as implicit text queries with the assistance of ChatGPT-4. For example, ``door\_handle'' yields ``open the door'', and ``table\_leg'' corresponds to ``parts that support the table''.

\begin{figure*}[th]
     \centering
     \vspace{-10px}
     \includegraphics[width=0.99\linewidth]{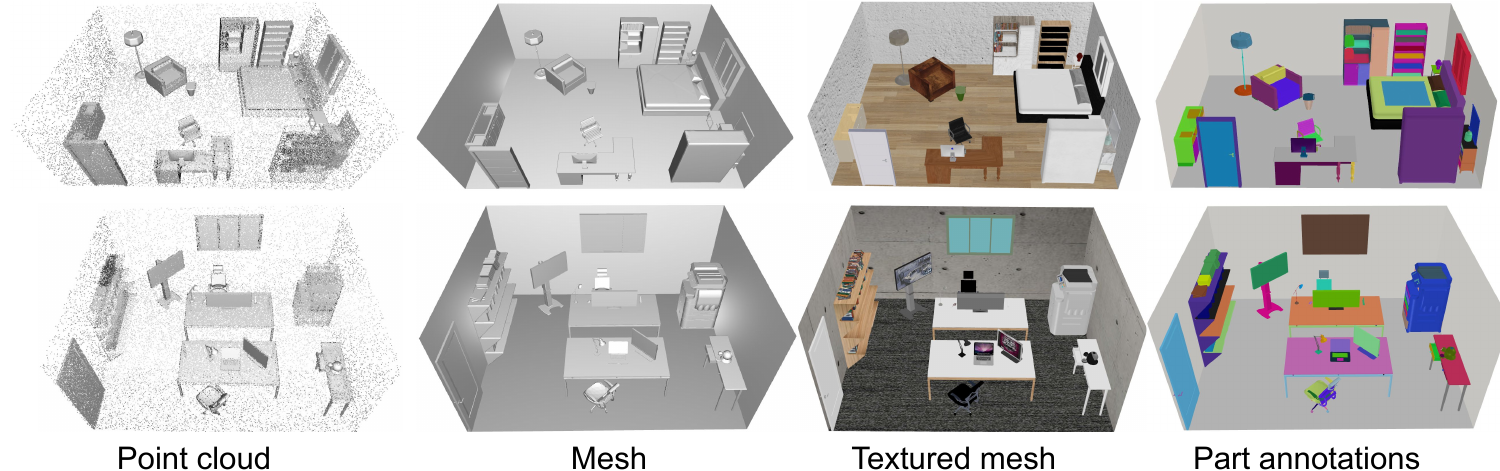}
	\caption{
        \textbf{3D-PU dataset scene examples.} 
	Each 3D scene includes a point cloud, mesh, textured mesh, and detailed part annotations for all objects.
 } 
	\label{fig:scene}
    \vspace{-5px}
\end{figure*}

Our guiding principle for creating 3D scenes is to maximize randomness and diversity, ensuring that each 3D scene is unique:
(i) Each scene is defined by a combination of a layout and multiple 3D objects. With 100 layouts and hundreds of thousands of 3D shapes, there are virtually endless possible combinations. Among our created 10,000 3D scenes, very few will share the same objects.
(ii) Layouts only specify the essential large objects and their placements, while small objects are randomly selected and positioned throughout the scene.
(iii) Even when the same object appears in different 3D scenes, it will have varied textures and colors.
(iv) When objects are placed into a 3D scene, they undergo slight random transformations, such as free-form deformation, scaling, and displacement.
These measures ensure that each 3D scene is entirely distinct from the others. Figure~\ref{fig:scene} showcases two 3D scenes from the 3D-PU dataset, each containing a point cloud, mesh, textured mesh, and detailed part annotations for all objects. The 10,000 3D scenes are divided into 9,000 for training, with 500 each allocated for validation and testing. Having established the annotated dataset we are now ready to highlight our 3D-part level segmentation task and method.

\vspace{-5px}

 \section{OpenPart3D}
\label{sec:method}
\vspace{-5px}

\begin{figure*}[th]
	\centering
	\includegraphics[width=0.9\linewidth]{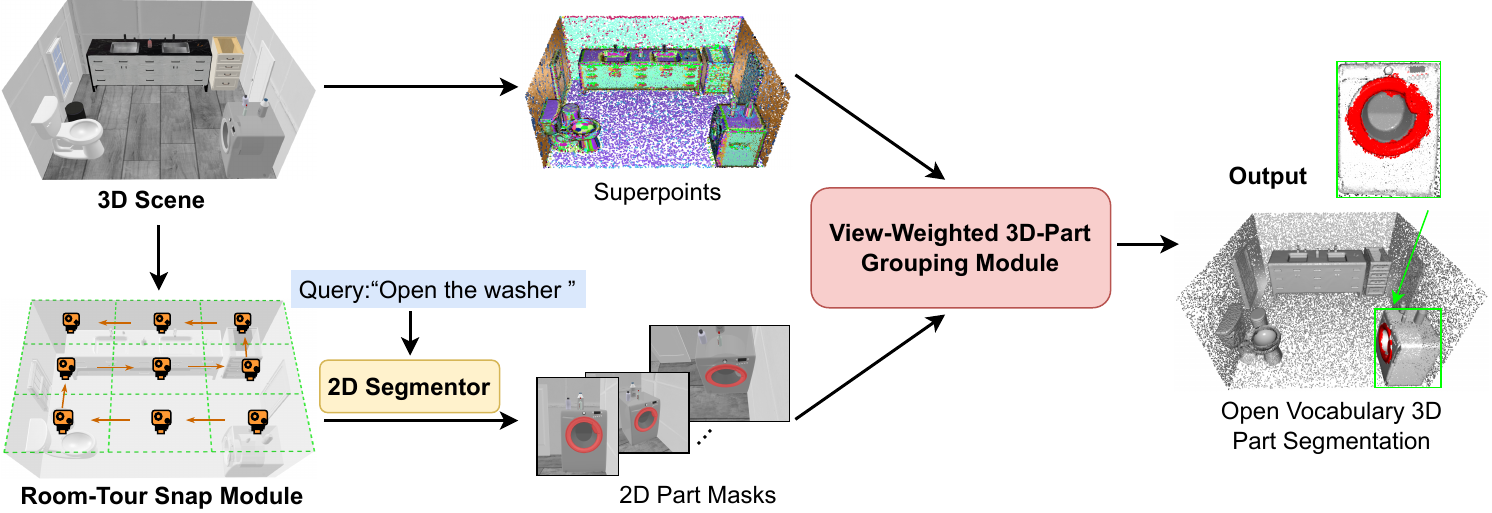}
	\caption{
        \textbf{Overview of OpenPart3D.} First, the~\textit{Room-Tour Snap Module} captures multiple view images of the 3D scene mesh by strategically positioning cameras at predefined locations with optimized poses. These images are then processed by a 2D open-vocabulary model to generate 2D part masks corresponding to the given text query. Subsequently, the~\textit{View-Weighted 3D-Part Grouping Module} integrates these 2D part masks across multiple views, assigning adaptive weights to each view, to extract geometrically consistent regions from the scene’s point cloud and aggregate them into precise 3D parts.
        \vspace{-8px}
} 
	\label{fig:method}
\end{figure*}

\noindent\textbf{Task definition.}
Given a scene-scale input point cloud $\mathcal{P}{=}\{( p_i, f_i)\}_{i=1}^{N}$ and a free-form text query $\mathcal{T}$, where $p_i \in \mathbb{R}^3$ represents the spatial coordinates of each point and $f_i$ denotes additional point features such as their RGB color, the objective is to predict fine-grained masks $\{ m_i \}_{i=1}^K$ for the 3D parts relevant to the text query $\mathcal{T}$. Here, $K$ denotes the number of parts in the scene that correspond to the specified text query $\mathcal{T}$.
%
%

\noindent\textbf{Method overview.} Our OpenPart3D model is illustrated in Figure~\ref{fig:method}.
First, the \textit{Room-Tour Snap Module} captures multiple view images of the 3D scene mesh by strategically positioning cameras at predefined locations with optimized poses, ensuring clear visibility of small objects and parts. These images are then processed by a 2D open-vocabulary model to generate 2D part masks corresponding to the given text query. Next, the \textit{View-Weighted 3D-Part Grouping Module} integrates these 2D part masks across multiple views, assigning adaptive weights to each view, to extract geometrically consistent regions from the scene’s point cloud and aggregates them into precise 3D parts. Detailed descriptions of each step are provided below.

\begin{wrapfigure}{r}{0.5\textwidth}
	\centering
        \vspace{-15px}
\includegraphics[width=0.49\textwidth]{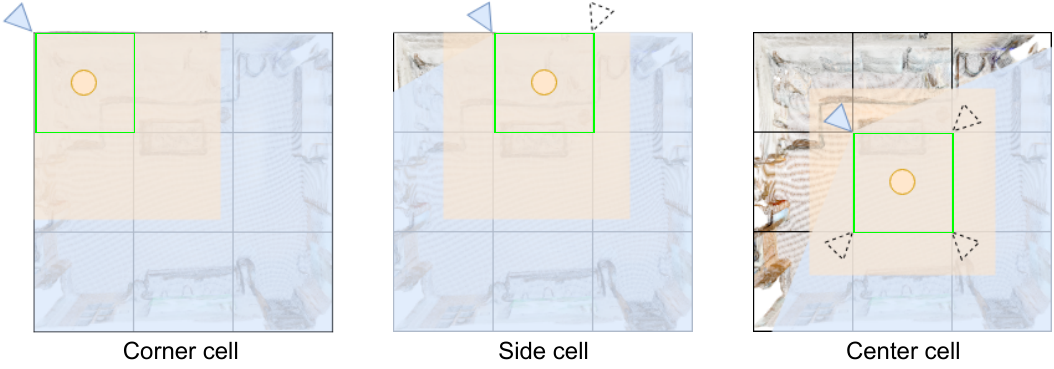}
	\caption{\textbf{Camera position and pose.} The orange circle \textcolor{orange}{\textbf{$\bigcirc$}} represents a camera positioned above the center of a grid cell, oriented toward the cell's centroid, while the blue triangle \textcolor{blue}{\textbf{$\bigtriangledown$}} indicates a camera oriented toward the farthest corner of the scene. Within the 3×3 grid of the scene, each cell contains a camera above its center, directed at its centroid. Additionally, corner cells include one camera pointing toward the farthest corner of the scene, side cells have two such cameras, and the center cell is equipped with four cameras oriented toward the farthest corners of the scene.    }
	\label{fig:camera}
    \vspace{-8px}
\end{wrapfigure}

\noindent\textbf{Room-tour snap.} Parts are typically small, making effective 2D part segmentation in subsequent steps challenging. To enhance the quality of part segmentation, it is crucial that each 3D object is not only present but also clearly visible in at least some of the projected view images. In other words, each object should be positioned near the virtual camera in at least some of the images. To address this, we propose the \textit{Room-Tour Snap Module}. As Figure~\ref{fig:camera} illustrates, the scene is divided into a 3x3 grid, with cameras strategically positioned above each grid cell. Each cell contains a camera placed above its center, oriented toward the centroid of the points within the cell. Additionally, corner cells include one camera aimed at the farthest corner of the scene, side cells have two such cameras, and the center cell is equipped with four cameras pointing toward the farthest corners of the scene.
In total, we capture 25 view images across all grid cells, ensuring comprehensive coverage of the scene and clear visibility of each object. 

\noindent\textbf{2D part segmentation.} 
Upon the projected view images, we apply a pretrained 2D open-vocabulary part segmentor, Florence2~\cite{xiao2024florence}, using the text query $\mathcal{T}$ as input. The network generates a set of 2D part masks corresponding to the text query. However, the performance of the vision-language model may be constrained by the specific images on which it was trained. To enhance the model's performance, we fine-tune the vision decoder of the VLM by leveraging ground-truth 2D part annotations projected from the 3D geometries, while freezing the other components of the VLM. Given the input $x$, which combines the projected images and the text query, and the ground-truth 2D masks $y$, the objective is formulated as follows:
\begin{equation}
    \mathcal{L} = -\sum_{i=1}^{|y|} \log P_{\theta}(y_i|y_{<i}, x),
\label{eq:ce_loss}
\end{equation}
where $\theta$ denotes the vision decoder parameters and $|y|$ represents the total number of target tokens.

\noindent\textbf{Superpoint generation.} 
In the pre-processing step, the scene points are grouped into geometrically homogeneous regions, referred to as superpoints~\cite{landrieu2018large,landrieu2017cut}. This results in a set of $S$ superpoints $\{ \hat{P}_i \}_{i=1}^{S} \in \{ 0,1 \}^{N \times S}$, where $\hat{P}_i$ represents a binary mask of the points. Superpoints serve as foundational elements for 3D parts, improving processing efficiency in subsequent stages.

\noindent\textbf{3D-part grouping.}
The \textit{View-Weighted 3D-Part Grouping Module} selects and merges the superpoints that correspond to the 2D part masks.
The possibility of each superpoint belonging to the \textit{foreground} class is voted by all 2D part segmentations across multiple projected view images that overlap with its 2D projection. Specifically, for a superpoint $\hat{P}_i$, its score $s_i$ is calculated as the weighted ratio of its visible points covered by all 2D part masks across all views:

\begin{equation}
s_i = \frac{\sum_v \sum_{p \in \hat{P}_i}\left[\text{VIS}_v(p) \right] \left[\text{INS}(p) \right]W_v}{\sum_v \sum_{p \in \hat{P}_i} \left[ \text{VIS}_v(p) \right]W_v},
\label{eq_score}
\end{equation}
where $[\cdot]$ represents the Iverson bracket (which evaluates to 1 if the predicate inside is true, and 0 if false); $\text{VIS}_v(p)$ indicates whether the 3D point $p$ is visible in view $v$; and $\text{INS}(p)$ indicates whether the projection of point $p$ in view $v$ is inside the part masks. 

Not all views contribute equally to the voting process, as certain views may provide more informative perspectives of the target instances. The weight $W_v = \{1,2,3\}$ is assigned to view $v$ based on the spatial proximity between its camera position and the corresponding part masks. A higher weight is assigned to views where the camera positions are closer to the part masks:
If the 2D masks and the camera are located within the same grid cell, $W_v =3$. If they are in neighboring grid cells, $W_v =2$. If they are in non-adjacent grid cells, $W_v =1$.

A superpoint is classified as \textit{foreground} if its score $s_i$ exceeds 0.5. The \textit{foreground} superpoints are then grouped to form 3D part masks, which are relevant to the text query.

\vspace{-5px}

 \section{Results}
\label{sec:result}
\vspace{-5px}

\subsection{Benchmarks and metrics}

\begin{table}
\vspace{-10px}
    \centering
    \caption{\textbf{Effect of fine-tuning} on 3D-PU and MultiScan datasets. Fine-tuning the vision decoder of our model on the synthetic 3D scene data from the 3D-PU dataset significantly enhances its 3D-part understanding capabilities, yielding improved performance on both 3D-PU and MultiScan datasets.}
    \resizebox{0.51\linewidth}{!}{
    \begin{tabular}{crrrr}
		\toprule
     & \multicolumn{2}{c}{\textbf{3D-PU}}&\multicolumn{2}{c}{\textbf{MultiScan}}\\
     \cmidrule(lr){2-3}
     \cmidrule(lr){4-5}
	 Fine-tuning  & $AP_{50}$ & $AP_{25}$& $AP_{50}$ & $AP_{25}$ \\
     \midrule
     \ding{55} & 10.4 & 21.4& 11.5 & 23.6\\
     \rowcolor{mygray}
    \ding{51} & 17.8 & 35.3& 13.7& 30.1\\
    & \textcolor{teal}{(+7.4)} &  \textcolor{teal}{(+13.9)}& \textcolor{teal}{(+2.2)} & \textcolor{teal}{(+6.5)} \\
     \bottomrule
     \end{tabular}
    }
    \label{tab:fintuning}
\end{table}

\begin{table*}[htbp]
    \caption{\textbf{Model ablation of key components} on 3D-PU dataset.}
    \label{tab:model_ablation}
    \centering
    \setlength{\tabcolsep}{3pt}
\vspace{-8px}
    {\small 
    
    \begin{tabular}{lr} 
        \begin{subtable}[t]{0.48\textwidth}
            
\centering
\setlength{\tabcolsep}{2pt}
\caption{\textbf{Comparing 2D-segmenters.} The open-vocabulary 2D model enables text query comprehension, with Florence2 showing the highest suitability for our approach. 
    }
{\begin{tabular}{lrr}
		\toprule
			 OpenPart3D & $AP_{50}$ & $AP_{25}$ \\
			\midrule
              w/ VLPart~\cite{sun2023going}   & 10.8 & 24.2\\
            w/ OpenSeeD~\cite{zhang2023simple}   & 13.1 & 29.3\\
            \rowcolor{mygray}
		   w/ Florence2~\cite{xiao2024florence}  & \textbf{17.8} & \textbf{35.3}\\
			\bottomrule
		\end{tabular}
\label{tab:2D_segmentor}
}

        \end{subtable}
        &
        \begin{subtable}[t]{0.48\textwidth}
            \centering
\caption{\textbf{Effect of varied view generation.} View images generated by our \textit{Room-Tour Snap Module} are effective in capturing small objects and parts, resulting in the best performance for part segmentation.
    }
{
    \begin{tabular}{lcrr}
		\toprule
			  View generation& Num of views & $AP_{50}$ & $AP_{25}$ \\
			\midrule
            Global snap &  16 & 14.4 & 30.1\\
            Multi-level  snap~\cite{huang2025openins3d} & 24 & 16.3 & 33.9\\
           \rowcolor{mygray}
           Room-tour snap (ours) & 25 & \textbf{17.8} & \textbf{35.3}\\
			\bottomrule
		\end{tabular}
    \label{tab:tour}
}

        \end{subtable}
    \end{tabular}

    \begin{tabular}{lr} 
        \begin{subtable}[t]{0.48\textwidth}
         \centering
\setlength{\tabcolsep}{2pt}
\caption{\textbf{Effect of view weight.} Placing more emphasis on views where the camera position is closer to the part masks can benefit part segmentation performance.}

{
\begin{tabular}{crr}
		\toprule
			 View weight  & $AP_{50}$ & $AP_{25}$ \\
			\midrule
              \ding{55} & 16.0 & 33.6\\
              \rowcolor{mygray}
              \ding{51} & \textbf{17.8} & \textbf{35.3}\\
			\bottomrule
		\end{tabular}
\label{tab:weight}
}

        \end{subtable}
        &
        \begin{subtable}[t]{0.48\textwidth}
        \centering
\setlength{\tabcolsep}{2pt}
\caption{\textbf{Influence of text query.} Direct queries result in better part segmentation because they are more straightforward and easier for the model to interpret. }

{\begin{tabular}{lrr}
		\toprule
			 Query & $AP_{50}$ & $AP_{25}$ \\
			\midrule
              Implicit   & 14.3 & 31.8\\
              Direct & 21.1 & 39.2\\
			\bottomrule
		\end{tabular}
\label{tab:query}
}

        \end{subtable}
    \end{tabular}

    } 
    \vspace{-8px}
\end{table*}

\noindent \textbf{Benchmarks.} We leverage the 3D-PU dataset as the benchmark for the task of open-vocabulary 3D part segmentation in scenes. We train the approach on the 3D-PU dataset training set and evaluate on the 3D-PU test set. For further evaluation, we also adapt the MultiScan dataset~\cite{mao2022multiscan}, which includes annotations for 5,129 parts across 5 part categories (static, door, drawer, window, lid) within 17 object categories in 117 scenes. We create ``object\_part'' labels based on each part’s category and associated object category, resulting in 47 unique labels, such as ``cabinet\_door'' and ``cabinet\_drawer.'' We use these ``object\_part'' labels as text queries.

\begin{wrapfigure}{r}{0.5\textwidth}
\vspace{-15px}
	\centering
\includegraphics[width=0.48\textwidth]{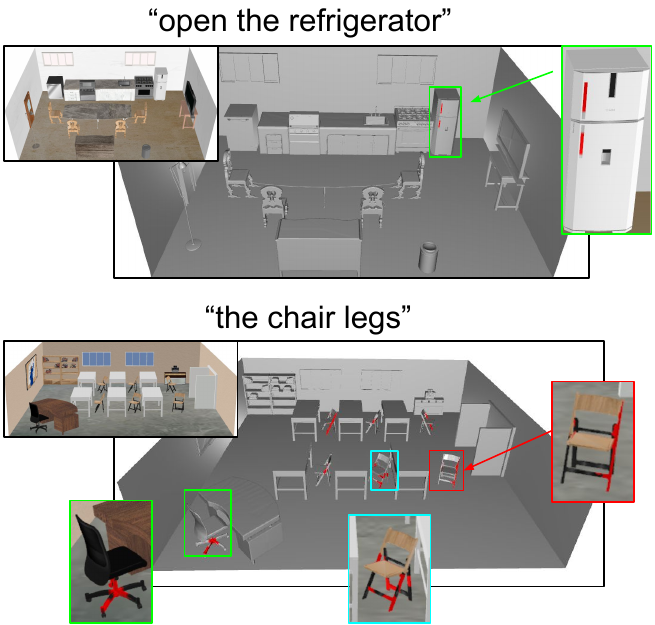}
	\caption{
 \textbf{Qualitative examples} from 3D-PU dataset. In the first scene, the small handle (in \textcolor{green}{ green box}) of the refrigerator is accurately segmented in response to the query ``open the refrigerator.'' In the second scene, most chair legs are identified, with only a few legs overlooked (in \textcolor{red}{red box}), demonstrating an overall strong performance.
 } 
	\label{fig:qualitative}
    \vspace{-30px}
\end{wrapfigure}

\noindent \textbf{Metrics.}  As primary metrics, we report the mean Average Precision (mAP) at Intersection over Union (IoU) thresholds of 0.25 and 0.5, referred to as $AP_{25}$ and $AP_{50}$. Additionally, we provide the AP, calculated as the average mAP across IoU thresholds from 0.5 to 0.95 in 0.05 increments.

\subsection{Ablation study}

\textbf{Effect of fine-tuning.} Table~\ref{tab:fintuning} compares the 3D part segmentation results with and without fine-tuning using the synthetic data from the 3D-PU dataset. Without fine-tuning, our approach already achieves reasonable performance on 3D-PU, indicating that the pre-trained vision-language model can effectively interpret the synthetic scenes in the 3D-PU dataset. Fine-tuning the vision decoder of our model on the 3D-PU training set enhances the part segmentation performance considerably on both 3D-PU and MultiScan datasets, demonstrating that the 3D scene data in the 3D-PU dataset can effectively improve the model's capability in 3D part understanding. In Figure~\ref{fig:qualitative}, we present two qualitative examples of 3D part segmentation from the 3D-PU dataset. 

\noindent \textbf{Comparing 2D-segmenters.} The 2D open-vocabulary segmenter is a key component of our approach, providing the ability to interpret text queries. In Table~\ref{tab:2D_segmentor}, we evaluate several open-vocabulary 2D segmentation models fine-tuned on the 3D-PU dataset. The results indicate the Florence2 model is the most suitable for our task.

\noindent \textbf{Effect of varied view generation.} The method of view generation plays a crucial role, as it directly impacts the ability of the view images to capture small objects and parts. Table~\ref{tab:tour} compares various view generation techniques on the 3D-PU dataset. The \textit{Global Snap} method moves the camera around the scene, always pointing directly toward the scene's center. The \textit{Multi-level Snap}~\cite{huang2025openins3d} captures view images at three different scales. The results demonstrate that the view images generated by our \textit{Room-Tour Snap} method lead to the best performance in part segmentation.

\noindent \textbf{Effect of view weight.} The \textit{View-weighted 3D-Part Grouping Module} prioritizes views where the camera position is closer to the 2D part masks. Table~\ref{tab:weight} illustrates that assigning different weights to views based on the distance between the camera position and the 2D part masks can enhance part segmentation performance.

\noindent \textbf{Influence of text query.} We use ``object\_part'' labels as direct queries and text descriptions as implicit queries. Table~\ref{tab:query} demonstrates that direct queries result in better part segmentation. This is understandable, as direct queries are more straightforward and easier for the model to interpret.

\begin{figure*}[th]
\vspace{-5px}
	\centering
	\includegraphics[width=0.99\linewidth]{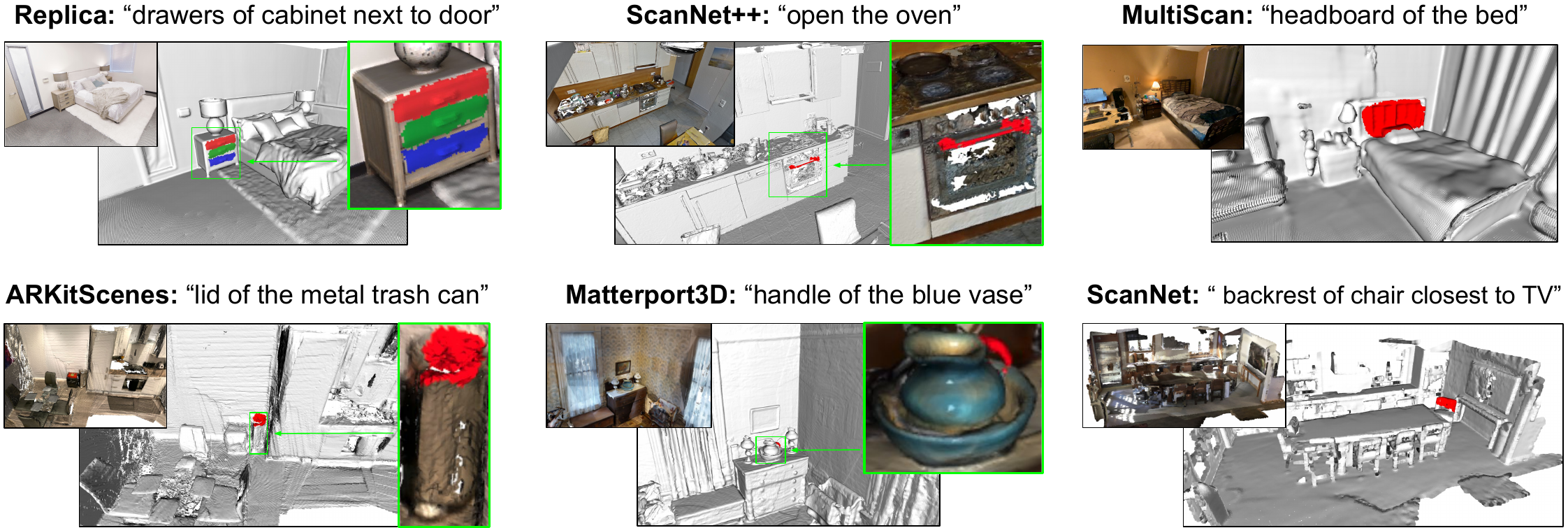}
	\caption{
        \textbf{Generalization to other datasets.} 
	Our approach can be applied to a wide range of common 3D datasets, including Replica, ScanNet++, MultiScan, ARKitScenes, Matterport3D, and ScanNet. We believe it is the first time 3D part segmentation works on these datasets.
 } 
	\label{fig:generalization}
    \vspace{-8px}
\end{figure*}

\noindent \textbf{Visualization examples on other datasets.} Our 3D part segmentation approach generalizes effectively across other 3D datasets. Figure~\ref{fig:generalization} showcases open-vocabulary 3D part segmentation visualizations on datasets such as Replica, ScanNet++, MultiScan, ARKitScene, Matterport3D, and ScanNet. We believe this marks the first application of 3D part segmentation on these datasets, highlighting the strong generalization and robustness of our approach.

\subsection{Comparison}

\begin{table}
\vspace{-10px}
	\centering
    \caption{\textbf{Open-vocabulary 3D part segmentation} on 3D-PU and MultiScan datasets. We constructed the baselines from the author-provided code. Our method achieves the best performance on both datasets, demonstrating its strong capability in 3D part segmentation.}
	\resizebox{0.8\columnwidth}{!}{%
\small
		\begin{tabular}{lrrrrrr}
			\toprule
				&\multicolumn{3}{c}{\textbf{3D-PU}}& \multicolumn{3}{c}{\textbf{MultiScan}}\\
				\cmidrule(lr){2-4} \cmidrule(lr){5-7}
			& $AP$ & $AP_{50}$ & $AP_{25}$ & $AP$ & $AP_{50}$ & $AP_{25}$\\
			\midrule
		OpenScene~\cite{peng2023openscene}  & 4.1 & 8.7 & 15.3 & 3.1 & 5.4  &  13.3\\
        OpenIns3D+PartDistill~\cite{huang2025openins3d,umam2024partdistill} & 4.7 & 9.8 & 19.5 & 3.9 & 7.3  & 15.2\\
          Search3D~\cite{takmaz2025search3d}  & 5.1 & 10.7 & 23.5 & 4.9 & 10.5  &  24.4\\
		\rowcolor{mygray}
		OpenPart3D  & \textbf{8.6} & \textbf{17.8} & \textbf{35.3} & \textbf{7.5} & \textbf{13.7} & \textbf{30.1}\\
			\bottomrule
		\end{tabular}%
	}
    
\label{tab:part_segmentation}
        \vspace{-8px}
\end{table}

\textbf{Open-vocabulary 3D-part segmentation in scenes.}
Since no existing baseline methods can be directly applied to the task and setting of open-vocabulary part segmentation in 3D scenes, we construct several strong baselines using the author-provided code. 
OpenScene~\cite{peng2023openscene} generates dense point-level features for 3D scenes, and we aggregate features of 3D parts based on cosine similarity to CLIP~\cite{radford2021learning} embeddings to respond to text queries. 
We construct the second baseline by concatenating OpenIns3D~\cite{huang2025openins3d} and PartDistill~\cite{umam2024partdistill}. So that OpenIns3D can localize the objects of interest in the scene according to the text query then PartDistill can further segment the parts within the objects of interest.
Search3D~\cite{takmaz2025search3d} constructing 2D object-part representations that correspond to 3D objects, relying on well-aligned 2D RGB-D images as input. We replace the well-aligned images by the generated images from the scene meshes and make Search3D suit our task and setting as the third strong baseline.
We train our method and the baseline methods on the 3D-PU training set and evaluate them on the 3D-PU test set and the MultiScan dataset. For fair comparison, we ensure consistent input and fine-tuning across all methods.
As shown in Table~\ref{tab:part_segmentation}, our method achieves the best performance on both datasets, demonstrating its superior 3D part segmentation capability.

\begin{wraptable}{r}{0.5\linewidth}
\vspace{-5px}
    \centering
    \caption{\textbf{Languge-guided functionality segmentation} on the SceneFun3D Dataset. Our method outperforms baseline models despite not utilizing well-aligned 2D images as input. }
    \scalebox{.95}{
    \begin{tabular}{lcrr}
\toprule

&\multicolumn{3}{c}{\textbf{SceneFun3D}}\\
\cmidrule(lr){2-4} 
& 2D input & $AP_{50}$ & $AP_{25}$ \\
\midrule
LERF~\cite{kerr2023lerf} & \ding{51} & 4.9 & 11.3\\
OpenMask3D-F~\cite{takmaz2023openmask3d} & \ding{51} & 8.0 & 17.5\\
\rowcolor{mygray}
OpenPart3D & \ding{55} & \textbf{9.3} & \textbf{21.7}\\
\bottomrule
\end{tabular}%
    }
\label{tab:functionality_segmentaion}
\end{wraptable}

\noindent\textbf{Language-guided functionality segmentation in 3D scenes.}
Language-guided functionality segmentation~\cite{delitzas2024scenefun3d} focuses on identifying functional, interactive elements in scenes based on a task description, such as ``open the door'' or ``turn on the ceiling light.'' These interactive elements are often small parts of objects, making our approach particularly well-suited for this task. We compare our method with LERF~\cite{kerr2023lerf} and OpenMask3D-F~\cite{takmaz2023openmask3d} on the SceneFun3D~\cite{delitzas2024scenefun3d} functionality segmentation benchmark. Notably, while both baselines rely on well-aligned 2D-3D pairs as input, our approach operates directly on 3D geometries alone. As shown in Table~\ref{tab:functionality_segmentaion}, our method outperforms the baselines even without aligned 2D images as input.

\vspace{-5px}


 \section{Discussion and conclusion}
\label{sec:conclusion}
\vspace{-5px}

\noindent\textbf{Discussion.}
\label{sec:discussion}
We introduce the 3D-PU dataset, the first large-scale 3D scene dataset with dense part-level annotations. While it encompasses a rich set of 85 object categories and 206 part categories, its category diversity is inherently constrained by the original object datasets from which it is derived, including PartNet, SAPIEN, and ShapeNetPart. Additionally, the data generation process is not fully automated, as the original part annotations are sourced from manual labeling efforts. In our construction, we adapt these part-level annotations from object understanding into scene understanding. And the scene layout generation also needs human validation.

Nevertheless, we believe 3D-PU represents a considerable advancement toward enabling comprehensive part-level annotation in 3D scenes. With the increasing importance of data curation and synthetic data generation in large-scale model research, 3D-PU strives to address a critical need in the 3D vision community. Despite being a synthetic dataset, 3D-PU demonstrates strong practical utility. When used for fine-tuning, it yields substantial performance improvements on real-world data, MultiScan dataset (see Table~\ref{tab:fintuning}). This effectiveness may be attributed to our fine-tuning strategy, where only the vision decoder is updated while other components remain frozen. This approach suggests that 3D-PU can play a crucial role in training robust and transferable models. Moreover, 3D-PU has the potential to serve as a valuable diagnostic dataset, analogous to CLEVR~\cite{johnson2017clevr}, tailored for the development of 3D foundation models in the near future.

\noindent\textbf{Conclusion.}
In this paper, we introduce the task of segmenting any 3D part in scenes based on a natural language description. To address data limitations, we propose a cost-effective approach for constructing 3D scene data with extensive, fine-grained part-level annotations. This method culminates in the creation of the first large-scale 3D scene dataset with detailed part annotations, termed the 3D-PU dataset. To tackle the challenges of open-vocabulary 3D part segmentation, we present OpenPart3D, a 3D-input-only framework. Our approach, evaluated on the 3D-PU dataset and the restructured MultiScan dataset, demonstrates superior performance over baseline methods in open-vocabulary 3D part segmentation, and showcases strong generalization across multiple common 3D scene datasets.

\vspace{0.5em}

\noindent\textbf{Acknowledgement.}
This work is financially supported by Qualcomm Technologies Inc., the University of Amsterdam, and the allowance Top consortia for Knowledge and Innovation
(TKIs) from the Netherlands Ministry of Economic Affairs and Climate Policy. 

{
    \clearpage
    \small
    \bibliographystyle{plainnat}
    \bibliography{sec/7_reference}
}

\end{document}